\documentclass[conference]{IEEEtran}
\usepackage{times}
\usepackage{graphics}
\usepackage{epsfig}
\usepackage{amsmath}
\usepackage{amssymb}
\usepackage{bm}
\usepackage{enumerate}
\usepackage[caption=false]{subfig}
\usepackage{diagbox}

\usepackage[numbers]{natbib}
\usepackage{multicol}
\usepackage[bookmarks=true]{hyperref}

\usepackage{anyfontsize}

\begin{document}

\title{Inverting Learned Dynamics Models for Aggressive Multirotor Control}

\author{\IEEEauthorblockN{Alexander Spitzer and Nathan Michael}
\IEEEauthorblockA{The Robotics Institute\\
Carnegie Mellon University\\
Pittsburgh, PA 15213, USA\\
\texttt{\{spitzer, nmichael\}@cmu.edu}}
}

\maketitle

\begin{abstract}
  We present a control strategy that applies inverse dynamics to a learned acceleration error model for accurate multirotor control input generation.
  This allows us to retain accurate trajectory and control input generation despite the presence of exogenous disturbances and modeling errors.
  Although accurate control input generation is traditionally possible when combined with parameter learning-based techniques, we propose a method that can do so while solving the relatively easier non-parametric model learning problem.
  We show that our technique is able to compensate for a larger class of model disturbances than traditional techniques can and we show reduced tracking error while following trajectories demanding accelerations of more than $7$ m/s$^2$ in multirotor simulation and hardware experiments.
\end{abstract}

\IEEEpeerreviewmaketitle

\section{INTRODUCTION}

\subsection{Motivation}

In the last several years, aerial robotics has seen a surge in popularity, largely due to the increasing viability of applications \cite{michael_collaborative_2012, kim_aerial_2013, cappo_online_2018}. Multirotors have been particularly well represented, due to their agility and versatility, and have additionally been a fruitful testbed for nonlinear controllers and trajectory generation strategies \cite{manchester_dirtrel_nodate, lee_geometric_2010, inaba_polynomial_2016, lee_feedback_2009}.

Computing precise control inputs for a dynamical system often requires accurate knowledge of its dynamics. \citet{nieuwstadt} showed how the concept of differential flatness can be used to generate control inputs that follow a given trajectory for \textit{differentially flat} systems.

For a multirotor, differential flatness can be used to compute the exact inputs required to follow a specified trajectory in $x$, $y$, $z$, and yaw (See \citet{mellinger}).
The computed control inputs are only accurate if the fixed dynamic model and its associated parameters, e.g. mass, inertia, etc., are correct. Often, this fixed dynamic model assumption fails and the estimated parameters are inaccurate. This results in suboptimal trajectory tracking performance.

One possible approach to alleviate this problem is to estimate the model parameters from vehicle trajectory data. This however, can be difficult, and is still suboptimal when the chosen parameterization cannot realize the true vehicle model. On the other hand, non-parametric error models are commonly used and relatively easy to learn but are not readily used in the differential flatness framework. In this work, we show how a non-parametric error model can be used to generate control inputs that follow a specified trajectory. We additionally provide an extension to the proposed approach that can deal with \textit{input-dependent} model errors via numerical optimization. We validate the control input generation strategy both in simulation and through experiments on a quadrotor.

\begin{figure}
  \begin{center}
    \includegraphics[width=8.7cm]{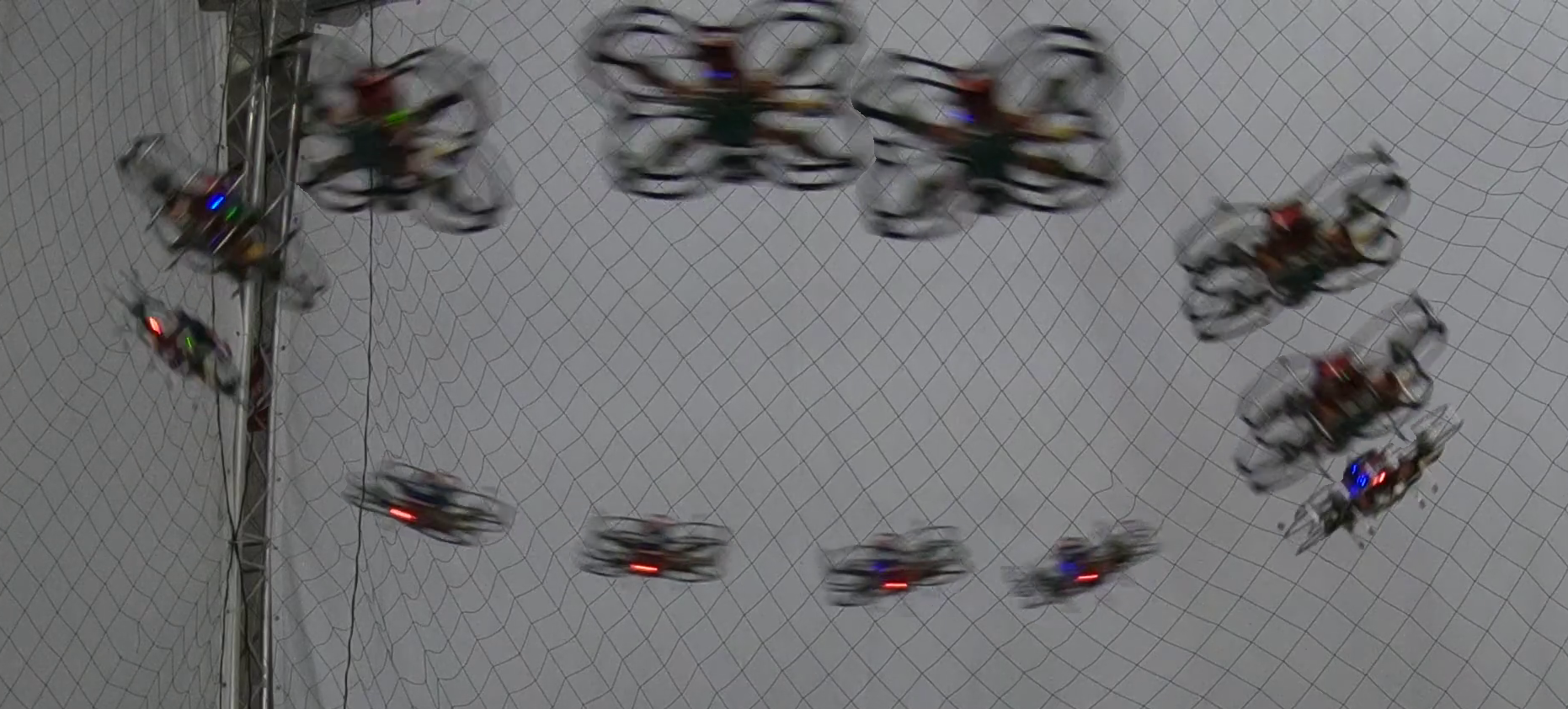} \\
    \vspace{0.3em}
    \includegraphics[width=8.7cm]{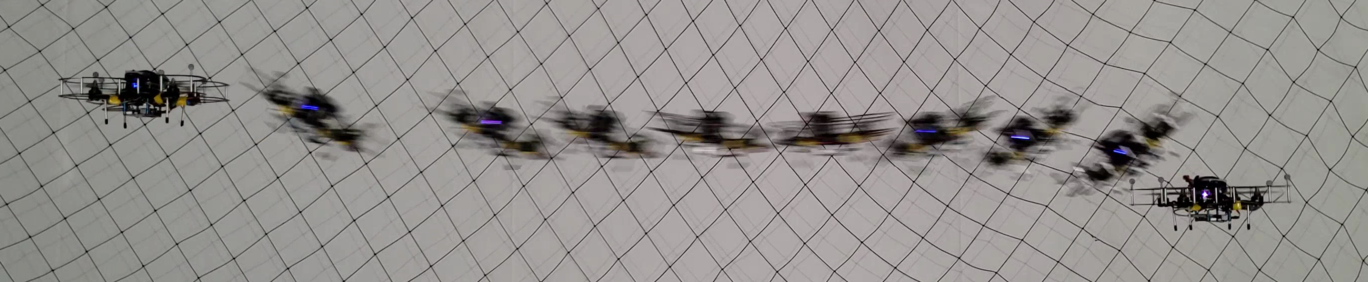}
    \includegraphics[width=9cm]{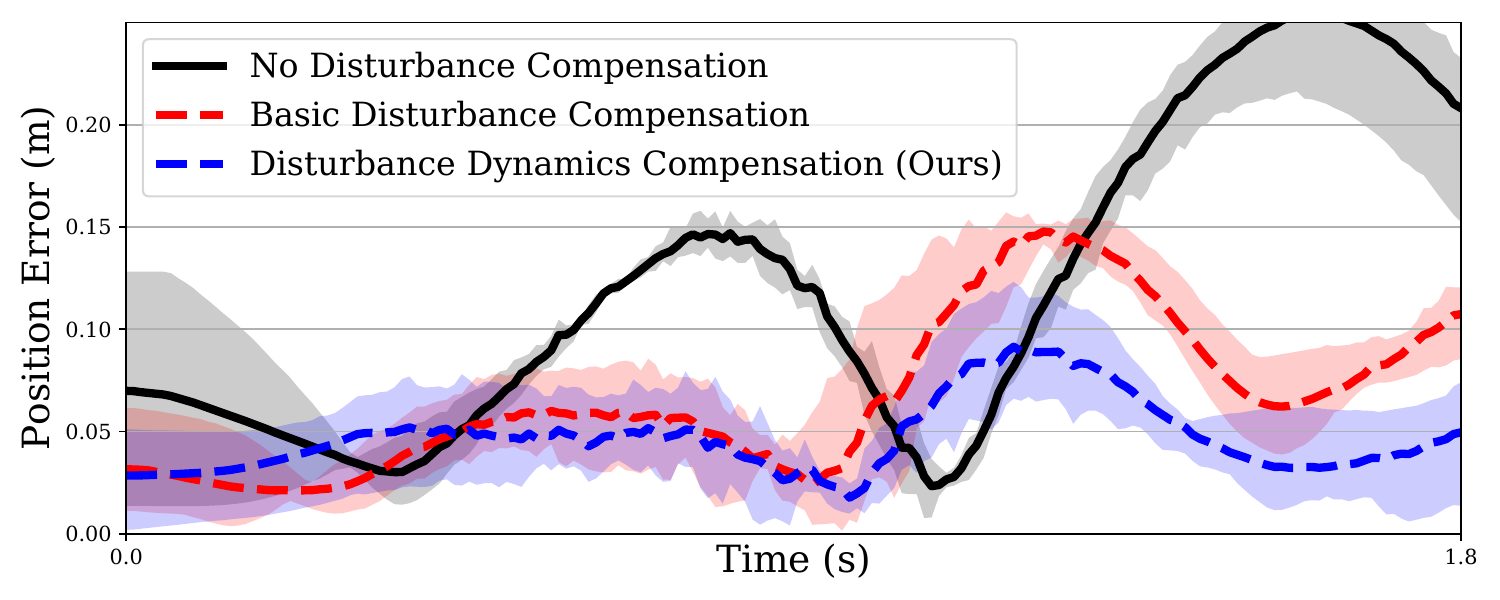}
  \end{center}
  \caption{Our experimental platform while executing an aggressive circle trajectory (top) and an aggressive line trajectory (middle) using the proposed control input generation strategy that is capable of compensating for dynamic and input-dependent acceleration disturbances (\textbf{FF5}). Our method substantially reduces tracking error along the aggressive line trajectory (bottom).}
  \label{fig:overlay}
\end{figure}

\subsection{Related Works}

Accurate and aggressive multirotor flight has been explored in \cite{tal, morrell, faessler17, mellinger} among others.
As for many other robotic platforms, accurate modeling has been shown to improve flight performance \cite{svacha, bangura}.
Traditional non-learning based modeling can be achieved via hand crafted experiments, calibration procedures, and computer-aided design \cite{mellinger_thesis}.
Since this requires significant manual effort and engineering hours, there have been many works exploring automatic parameter estimation methods \cite{burri_mle, burri_gif} and non-parametric model learning methods \cite{li, desaraju_thesis} for multirotor control.
In this work, we focus on non-parametric model learning methods, since parameter learning methods can be limited in their accuracy by the choice of parameterization \cite{nguyen-tuong_model_2011}.
There has also been work on learning control input corrections for aggressive flight without learning a dynamical model \cite{lupashin}.
These methods are not a focus of this paper since they can typically only be applied while executing the trained trajectories or reference quantities.
A learned dynamical model can be applied to any trajectory or reference.

Non-parametric model learning methods for robot control have been employed in \cite{florez, schaal, abbeel, mckinnon}.
Model learning performed in real-time incrementally has been studied in \cite{grollman, gijsberts11, balakrishnan}.
\citet{florez} use Locally Weighted Projection Regression (LWPR \cite{vijayakumar}) while \citet{gijsberts11} use Random Fourier Features \cite{rahimi}, which was extended to Incremental Sparse Spectrum Gaussian Process Regression (ISSGPR), a Bayesian regression formulation, in \citet{gijsberts13}.
\citet{droniou} evaluated LWPR and ISSGPR for the purposes of robot control and found ISSGPR to perform better.
In this work, we use linear regression, but our approach can use any model learning strategy.

Once an accurate dynamical system model is known, a Model Predictive Control (MPC) strategy can be used to optimize a desired cost function, subject to the dynamics \cite{desaraju_thesis, manchester_dirtrel_nodate, li_todorov, aswani}.
These approaches often make approximations to ensure real time feasibility \cite{desaraju_thesis, aswani}.
Furthermore, \citet{desaraju_thesis} does not perform full inverse dynamics on the disturbance, which can lead to suboptimal performance while tracking aggressive trajectories.

The differential flatness property of multirotors has been widely exploited for accurate trajectory tracking \cite{faessler, hehn, mellinger, ferrin, rivera, morrell}.
Differential flatness of the multirotor subject to linear drag was shown in \citet{faessler}.
This extends the applicability of the approach to a limited family of disturbances.
\citet{faessler} do not address the issue of nonlinear disturbances as a function of state and/or control input in the flatness computations.
Issues arising from singularities, commonly encountered during aggressive flight, were discussed and mitigated in \citet{morrell}, increasing the robustness of the differential flatness approach.

Although control inputs computed using the differential flatness framework will automatically take into account dynamical model parameter changes, such as mass, inertia, etc., it is not clear how to incorporate non-parametric model corrections.
In this work, we build on the differential flatness formulation by extending it to compensate for learned non-parametric dynamic model disturbances.
Our approach can compensate for arbitrary disturbances that are a function of vehicle position and velocity, as well as control input dependent disturbances that are a function of vehicle orientation and thrust. This increases the applicability of the approach to a much wider range of realistic flight conditions.

\subsection{Notation}

Lowercase letters such as $u$ and $z$ are scalars in $\mathbb{R}$.
Boldface lowercase letters such as $\bm{u}$ and $\bm{z}$ are vectors.
$\bm I_n$ is the $n$ by $n$ identity matrix.
$\dot{x}$ denotes the total time derivative of $x$.
$m$ is mass, $g$ is the gravitational constant and $I$ is inertia.
All functions in this paper are assumed to have continuous second derivatives everywhere and thus all second partial derivatives are symmetric, i.e. $\frac{\partial f}{\partial x \partial y} = \frac{\partial f}{\partial y \partial x}$.
Unless otherwise indicated, all vector quantities are expressed in a fixed reference frame.

\section{METHOD}

In this section, we first introduce the problem statement in Part \textit{A}.
Part \textit{B} details our approach for compensating for dynamic disturbances that can be a function of vehicle position, vehicle velocity, or other quantities that are independent of the applied control inputs.
Part \textit{C} extends the approach to compensate for disturbances that are \textit{input-dependent} and can be a function of e.g. the applied vehicle thrust or vehicle orientation.
Finally, Part \textit{D} describes the model learning approach.

\subsection{Problem Statement}

Assume we are given a desired position over time, $\bm{x}_d(t) \in \mathbb{R}^3$, along with its first four time derivatives, the velocity, acceleration, jerk, and snap: $\bm{v}_d(t)$, $\bm{a}_d(t)$, $\bm{j}_d(t)$, $\bm{s}_d(t)$.

Equation~\eqref{eq:accel_model} shows a typical acceleration model of a multirotor, where the commanded acceleration is aligned with the body $z$-axis.
\begin{equation}
  \bm{a} = u\bm{z} + \bm{g} + \bm{f}_e(\bm{\eta}, u)
  \label{eq:accel_model}
\end{equation}

Here $u \in \mathbb{R}$ is the commanded body acceleration, $\bm{z} \in \mathbb{R}^3$ is the body $z$-axis ($||\bm{z}|| = 1$), $\bm{g} = \begin{bmatrix} 0 & 0 & -g \end{bmatrix}^\top$ is the gravity vector, and $\bm{f}_e \in \mathbb{R}^3$ is an additive acceleration error model that can, in general, be a function of both vehicle state $\bm{\eta}$ and control input $u$.

The objective is to compute the body acceleration $u$, body $z$-axis $\bm{z}$, angular velocity $\bm\omega$, and angular acceleration $\dot{\bm \omega}$ such that integrating $\dot{\bm \omega}$ forwards in time twice results in an orientation with $\bm{z}$ as the $z$-axis and that the vehicle acceleration, which is a function of $u\bm{z}$, equals the desired vehicle acceleration $\bm{a}_d(t)$.
This will ensure that the vehicle follows the specified trajectory $\bm{x}_d(t)$.
Note that while $\bm z$ and $\bm\omega$ are not true control inputs to the system, they are necessary as feedforward references to the attitude feedback controller.
Once the body acceleration $u$ and angular acceleration $\dot{\bm{\omega}}$ are computed, they are multiplied by mass and inertia and used as the feedforward force and torque in the position and attitude feedback controllers respectively.

For simplicity, we will assume that the yaw of the vehicle is always zero, but all of the methods presented are applicable while following yaw trajectories as well.

\subsection{Input-independent error compensation}

The simplest version of our control input generation strategy assumes that the disturbance model $\bm{f}_e$ is a function of the vehicle position and velocity only: $\bm{f}_e(\bm{x}, \dot{\bm{x}})$.
In this case, the desired acceleration vector can be computed directly, as shown in \eqref{eq:accel_in}.

\begin{equation}
u\bm{z} = \bm{a}_d - \bm{g} - \bm{f}_e(\bm{x}, \dot{\bm{x}})
  \label{eq:accel_in}
\end{equation}

Since $\bm{z}$ must be of unit length, both $u$ and $\bm{z}$ can be computed from $u\bm{z}$ by computing the magnitude and normalizing.

The angular velocity and angular acceleration are found by first computing the first and second time derivatives of $\bm{z}$.

Differentiating \eqref{eq:accel_in} in time results in

\begin{equation}
  \dot u \bm{z} + u \dot{\bm{z}} =
    \bm{j}_d
    - \frac{\partial \bm{f}_e}{\partial \bm{x}} \dot{\bm{x}}
    - \frac{\partial \bm{f}_e}{\partial \dot{\bm{x}}} \ddot{\bm{x}} = \bm{j}_d^{\text{eff}}
    \label{eq:accel_in_d}
\end{equation}

Since $\bm{z}$ is of unit length, and it must remain so, it must be perpendicular to $\dot{\bm z}$.
Thus taking a dot product of \eqref{eq:accel_in_d} with $\bm{z}$ allows us to find $\dot u$.

\begin{equation}
  \dot u =
  {\bm{j}_d^{\text{eff}}}^\top\bm{z}
\end{equation}

Inserting $\dot u$ into \eqref{eq:accel_in_d} gives us $\dot{\bm z}$.

\begin{equation}
  \dot{\bm z} =
    \frac{1}{u}
    \left(
      \bm{j}_d^{\text{eff}}
      - \left(
        {\bm{j}_d^{\text{eff}}}^\top\bm{z}
        \right)\bm{z}
    \right)
\end{equation}

The body angular velocity can be extracted from $\dot{\bm z}$ by first defining the body $x$ and body $y$-axes using a desired vehicle yaw, then projecting $\dot{\bm z}$ onto those axes. See \cite{mellinger} for the details.

To find $\ddot{\bm z}$, we differentiate \eqref{eq:accel_in_d}.

\begin{equation}
  \ddot u \bm{z} + 2 \dot u \dot{\bm z} + u \ddot {\bm z} = \bm{s}_d - \ddot{\bm f}_e(\bm x, \dot{\bm x}) = \bm{s}_d^{\text{eff}}
  \label{eq:accel_in_dd}
\end{equation}

The second time derivative of the learned disturbance $\bm{f}_e$ is shown in \eqref{eq:fedd}. Note that the second partial derivative of the error model with respect to its vector inputs is a 3rd order tensor.

\begin{align}
  \ddot{\bm{f}}_e =
    &\left(
      \frac{\partial ^2 \bm{f}_e}{\partial \bm x ^2} \dot{\bm x}
      +
      \frac{\partial ^2 \bm{f}_e}{\partial \bm x \dot{\bm x}} \ddot{\bm x}
    \right)
    \dot{\bm x}
    +
    \frac{\partial \bm{f}_e}{\partial \bm x} \ddot{\bm x}
    + \nonumber \\
    &\left(
      \frac{\partial ^2 \bm{f}_e}{\partial \dot{\bm x}\bm x} \dot{\bm x}
      +
      \frac{\partial ^2 \bm{f}_e}{\partial \dot{\bm x}^2} \ddot{\bm x}
    \right)
    \ddot{\bm x}
    +
    \frac{\partial \bm{f}_e}{\partial \dot{\bm x}} \dddot{\bm x}
  \label{eq:fedd}
\end{align}

Noting that differentiating $\bm z^\top \dot{\bm z} = 0$ implies $\bm z ^\top \ddot{\bm z} = - \dot{\bm z} ^\top \dot{\bm z}$ and again taking a dot product with $\bm z$, we can compute $\ddot u$.

\begin{equation}
\ddot u = {\bm{s}_d^{\text{eff}}}^\top\bm z + u\dot{\bm z}^\top \dot{\bm z}
\end{equation}

Inserting $\ddot u$ into \eqref{eq:accel_in_dd} gives us $\ddot{\bm z}$.

\begin{equation}
  \ddot{\bm z} = \frac{1}{u}\left(
    \bm{s}_d^{\text{eff}} - \ddot u \bm z - 2\dot u \dot{\bm z}
  \right)
\end{equation}

To compute the body angular acceleration from $\ddot{\bm z}$, we note that $\ddot{\bm z} = \dot{\bm \omega} \times \bm z + \bm \omega \times \dot{\bm z}$ and proceed as before for the angular velocity, by projecting $\ddot{\bm z} - \bm \omega \times \dot{\bm z}$ onto the body $x$ and $y$-axes.

Note that the above equations for $\dot{\bm z}$ and $\ddot{\bm z}$ are similar to those derived in \cite{mellinger} with the difference that here, the first and second derivatives of the learned dynamics model are incorporated. In this way, the control inputs generated \textit{anticipate} changes in the disturbance.

One practical issue that arises is that the vehicle acceleration, $\ddot{\bm x}$, and jerk, $\dddot{\bm x}$, are not readily available during operation. Computing them from odometry by taking finite-differences will introduce noise. To alleviate this in our experiments, we use the acceleration and jerk demanded by the trajectory, which are good approximations of the true vehicle acceleration and jerk when tracking error is low.

\subsection{Input-dependent error compensation}

In many cases, additive dynamics model errors are a function of the applied control input and vehicle orientation, in addition to the vehicle position and velocity. For example, if the mass of the vehicle is not accurately known (or alternatively, the actuators are not properly modeled), the disturbance will be a linear function of the applied acceleration. The input-dependent acceleration model is shown in \eqref{eq:accel_de}.

\begin{equation} \label{eq:accel_de}
  u\bm{z} = \bm{a}_d - \bm{g} - \bm{f}_e(\bm{\eta}, \bm{u})
\end{equation}

Here, $\bm{\eta} = \begin{bmatrix} \bm x & \dot{\bm x} \end{bmatrix}^\top$ contains the vehicle position and velocity and $\bm u = u \bm z$.

Without assuming a particular form for the additive error term $\bm{f}_e$, it is not possible to solve for the required acceleration and orientation analytically. We must resort to solving the problem numerically. Interestingly however, once a solution for the acceleration and orientation is found, the rest of the control inputs can be found analytically in a method similar to the input-independent case described above.

We first rewrite the acceleration model as the functional equation $\bm f(\bm u, t) = 0$ that is only a function of $\bm u$ and time.
We compute the time derivative of $\bm u$ by taking a derivative of the above equation and solving the resulting linear system.

\begin{align} \label{eq:f_gen}
  \dot{\bm f}(\bm u, t) &= \frac{\partial \bm f}{\partial \bm u} \dot{\bm u} + \frac{\partial \bm f}{\partial t} = 0 \\
  \dot{\bm u} &=
    - \left( \frac{\partial \bm f}{\partial \bm u}\right)^{-1}
    \frac{\partial \bm f}{\partial t}
\end{align}

For our acceleration model, $\bm{f}(\bm u) = \bm u + \bm g + \bm{f}_e(\bm{\eta}, \bm u) - \bm{a}_d$. The necessary derivatives are shown in \eqref{eq:dfdu} and \eqref{eq:dfdt}.

\begin{equation}
  \label{eq:dfdu}
  \frac{\partial \bm f}{\partial \bm u} = \bm{I}_3 +
  \frac{\partial \bm f_e}{\partial \bm u}
\end{equation}

\begin{equation}
  \label{eq:dfdt}
  \frac{\partial \bm f}{\partial t} =
    \frac{\partial \bm f_e}{\partial \bm \eta} \dot{\bm \eta}
    - \bm{j}_d
\end{equation}

To find $\ddot{\bm u}$, we take a derivative of \eqref{eq:f_gen} and again solve the resulting linear system.

\begin{align}
  \ddot{\bm f}(\bm u, t)
  &=
  \left(
    \frac{\partial^2\bm f}{\partial \bm u^2}
      \dot{\bm u}
    +
    2
    \frac{\partial^2\bm f}{\partial \bm u \partial t}
  \right)
  \dot{\bm u}
  +
  \frac{\partial \bm f}{\partial \bm u}
    \ddot{\bm u}
  +
  \frac{\partial^2\bm f}{\partial t^2} \\
  \ddot{\bm u}
  &=
    \left( \frac{\partial \bm f}{\partial \bm u}\right)^{-1}
    \left(
      -
      \left(
        \frac{\partial^2\bm f}{\partial \bm u^2}
          \dot{\bm u}
        +
        2
        \frac{\partial^2\bm f}{\partial \bm u \partial t}
      \right)
      \dot{\bm u}
    -
    \frac{\partial^2\bm f}{\partial t^2}
    \right)
\end{align}

The necessary derivatives for our acceleration model are shown in \eqref{eq:d2fdu2}, \eqref{eq:d2fdudt}, and \eqref{eq:d2fdt2}.

\begin{align} \label{eq:d2fdu2}
  \frac{\partial^2\bm f}{\partial \bm u^2} =
  \frac{\partial^2\bm f_e}{\partial \bm u^2}
\end{align}
\begin{align} \label{eq:d2fdudt}
  \frac{\partial^2\bm f}{\partial \bm u \partial t} =
  \frac{\partial^2\bm f_e}{\partial \bm u \partial \bm \eta} \dot{\bm \eta}
\end{align}
\begin{align} \label{eq:d2fdt2}
  \frac{\partial^2\bm f}{\partial t^2} =
  \left(
    \frac{\partial^2 \bm f_e}{\partial \bm \eta ^2} \dot{\bm \eta}
  \right) \dot{\bm \eta}
  +
  \frac{\partial \bm f_e}{\partial \bm \eta}\ddot{\bm \eta}
  -
  \bm s_d
\end{align}

To compute $\dot{\bm z}$ from $\dot{\bm u}$, we take a derivative of $\bm u = u \bm z$ and proceed as before, by projecting onto $\bm z$ and solving first for $\dot u$.

\begin{align}
  \dot{\bm u} &= \dot u \bm z + u \dot{\bm z} \label{eq:du} \\
  \dot u &= \dot{\bm u}^\top \bm z \\
  \dot{\bm z} &= \frac1u \left(\dot{\bm u} - \dot u \bm z \right)
\end{align}

To compute $\ddot{\bm z}$ from $\ddot{\bm u}$, and the angular velocity and angular acceleration, we follow the same approach as for the input-independent case.

It should be noted that this approach requires the existence of a solution to $\frac{\partial \bm f}{\partial \bm u} \dot{\bm u} = \frac{\partial \bm f}{\partial t}$ and the analogous equation for $\ddot{\bm u}$. Solutions will only fail to exist when the estimated disturbance model is strong enough to completely negate the acceleration imparted by $\bm u$. This may be a concern when learning a model from data, but in practice has not occurred in our experiments.

\subsection{Model Learning}

To estimate $\bm{f}_e$ from vehicle trajectory data, we fit a model to differences between the observed and the predicted acceleration at every time step.
The observed acceleration is computed using finite-differences of the estimated vehicle velocity while the predicted acceleration is $u\bm z + \bm g$.

In principle, any regressive model whose derivatives are available can be used.

\section{EXPERIMENTS}

We first evaluate the proposed approach on a simulated 2D multirotor that is subjected to a series of input-independent and input-dependent disturbances. We then evaluate how the approach reduces tracking error on a quadrotor executing aggressive trajectories.

\subsection{Simulation}

The 2D planar multirotor captures many of the important dynamics present in the 3D multirotor. Namely, orientation and acceleration are coupled. In fact, the motion of a 3D multirotor moving in a vertical plane, e.g. in a straight line trajectory, can essentially be described with the 2D multirotor. As such, we believe a planar simulation is an appropriate testbed for our method.

\begin{figure}[h]
  \begin{center}
  \includegraphics{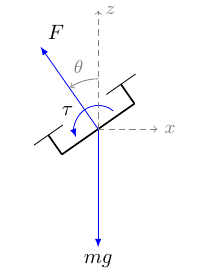}
  \end{center}
  \caption{Force diagram of the 2D multirotor used in the simulation experiment. $F$ and $\tau$ are the control inputs.}
\label{2dquad}
\end{figure}

The 2D multirotor force model is shown in Fig.~\ref{2dquad}.
The dynamics are shown in \eqref{dyn2d1} -- \eqref{dyn2d3}, where $F$ is the applied body force and $\tau$ is the applied body acceleration. The mass, $m$, was set to $4.19$ kg, gravity $g$ to $10.18$ m/s$^2$, and inertia $I$ to $0.123$ kg-m$^2$.

\begin{align}
  m \ddot x &= -F \sin(\theta) \label{dyn2d1} \\
  m \ddot z &= F \cos(\theta) \label{dyn2d2} - mg \\
  I \ddot \theta &= \tau \label{dyn2d3}
\end{align}

We subject the simulated multirotor to disturbances selected from Table~\ref{tab:dists}. Disturbance 1 is constant and emulates a fixed force field in the $x$ direction, e.g. due to wind. It is not input-dependent and is not dynamic since it does not change along with the vehicle state. Disturbance 2 is velocity dependent and emulates drag in the $x$ direction. Disturbance 3 depends on the vehicle angle and is thus input-dependent. Disturbance 4 is velocity dependent and emulates drag in the $z$ direction. Disturbance 5 is a mass perturbation that adds a disturbance linear in the applied acceleration, which makes it input-dependent.

\begin{table}[h]
\caption{Disturbances used in the 2D multirotor simulation experiment.}
\label{tab:dists}
\begin{center}
\begin{tabular}{|c|c|c|c|}
  \hline \textbf{No.} & \textbf{Effect} & \textbf{Input-dependent?} & \textbf{Dynamic?} \\
  \hline 1 & $\ddot x$ -= $4.1$ & no & no \\
  \hline 2 & $\ddot x$ -= $3.1\dot x$ & no & yes \\
  \hline 3 & $\ddot x$ += $1.4\sin(\theta)$  & yes & yes \\
  \hline 4 & $\ddot z$ -= $3.1\dot z$ & no & yes \\
  \hline 5 & $m$ += $2$ & yes & yes\\
  \hline
\end{tabular}
\end{center}
\end{table}

The vehicle is given a desired trajectory that takes it from $x = 0$, $z = 0$, to $x = 1$, $z = 1$ in one second.
The trajectories in $x$ and $z$ are both 7th-order polynomials that have the velocity, acceleration, and jerk equal to zero at each of their endpoints.
This ensures that the trajectory starts and ends with the vehicle at rest, at an angle of zero, and with an angular velocity of zero.
When Disturbance 1 is in effect, the vehicle's angle is initialized such that maintaining zero acceleration in $z$ also maintains zero acceleration in $x$. This ensures that the trajectory can be perfectly followed with correct control inputs despite the constant acceleration disturbance in $x$. In all other cases, the vehicle state starts at 0.

We show $x$ and $z$ tracking error for the following feedforward input generation strategies with and without feedback.

\begin{itemize}
  \item[\textbf{FF1})] No disturbance learning
  \item[\textbf{FF2})] Basic disturbance compensation (no disturbance dynamics)
  \item[\textbf{FF3})] Disturbance compensation w/ numerical optimization
  \item[\textbf{FF4})] Dist. comp. w/ disturbance dynamics (ours)
  \item[\textbf{FF5})] Dist. comp. w/ num. opt. and disturbance dynamics (ours)
\end{itemize}

\textbf{FF1} uses the feedforward generation strategy as presented in \cite{mellinger} and does not do any regression for disturbance learning.
\textbf{FF2} and \textbf{FF3} do not consider the dynamics of the disturbance; they compute the angular velocity and angular acceleration feedforward terms as in \cite{mellinger} while incorporating the learned disturbance in the acceleration model, \eqref{eq:accel_model}.
\textbf{FF4} is the proposed approach that deals with input-independent disturbances while \textbf{FF5} is the proposed approach that deals with input-dependent disturbances.

In this experiment, \textbf{FF3} and \textbf{FF5} solve \eqref{eq:accel_de} numerically using the modified Powell method root finder in SciPy \cite{scipy}. The initial guess for the optimization is the solution from the previous timestep.

Position and angle feedback is provided by PD controllers with gains of 10 on position and velocity errors, 300 on angle errors, and 30 on angular velocity errors.
The position PD controller output is added to the desired acceleration and the angle PD controller output is added to the desired angular acceleration.

In all simulation experiments, the feature vector used for linear regression of model errors is shown in \eqref{eq:2dfeat}.
The features were hand selected to appropriately model the disturbances in Table~\ref{tab:dists}.

\begin{equation}
  \bm{\phi}(x, z, \theta, \dot x, \dot z, F) = \begin{bmatrix} x, z, \dot x, \dot z, \sin(\theta), F\sin(\theta), F\cos(\theta), 1 \end{bmatrix}^\top
  \label{eq:2dfeat}
\end{equation}

The learned model is thus

\begin{equation}
  \bm{f}_e(\bm \eta, \bm u) = \bm w ^\top \bm\phi(\bm \eta, \bm u)
\end{equation}

$\bm w$ is the result of regressing the projected input data $\bm \phi$ to the observed acceleration errors and minimizing least squared error.
In this experiment, $\bm w$ is recomputed after every trajectory execution using data from all past executions. Results reported are on the 3rd run, since we found that only two regression steps were needed to converge to an accurate enough model. This is not surprising, as in this simulation there is no noise and the features used can appropriately reproduce the applied disturbances.

Each control configuration is subjected to the following set of disturbance combinations.

\begin{itemize}
  \item[{\textbf{A}})] Disturbance 1
  \item[{\textbf{B}})] Disturbances 1, 2, and 4
  \item[{\textbf{C}})] Disturbances 3 and 5
  \item[{\textbf{D}})] Disturbances 1, 2, 3, 4, and 5
\end{itemize}

\begin{figure*}[t!]
  \centering
  \begingroup
  \captionsetup[subfigure]{width=8cm}
  \subfloat[Errors for disturbance set \textbf{A}, containing only a constant disturbance. All strategies that compensate for the disturbance perform well.\label{fig:res_a}]{\includegraphics[width=8.3cm]{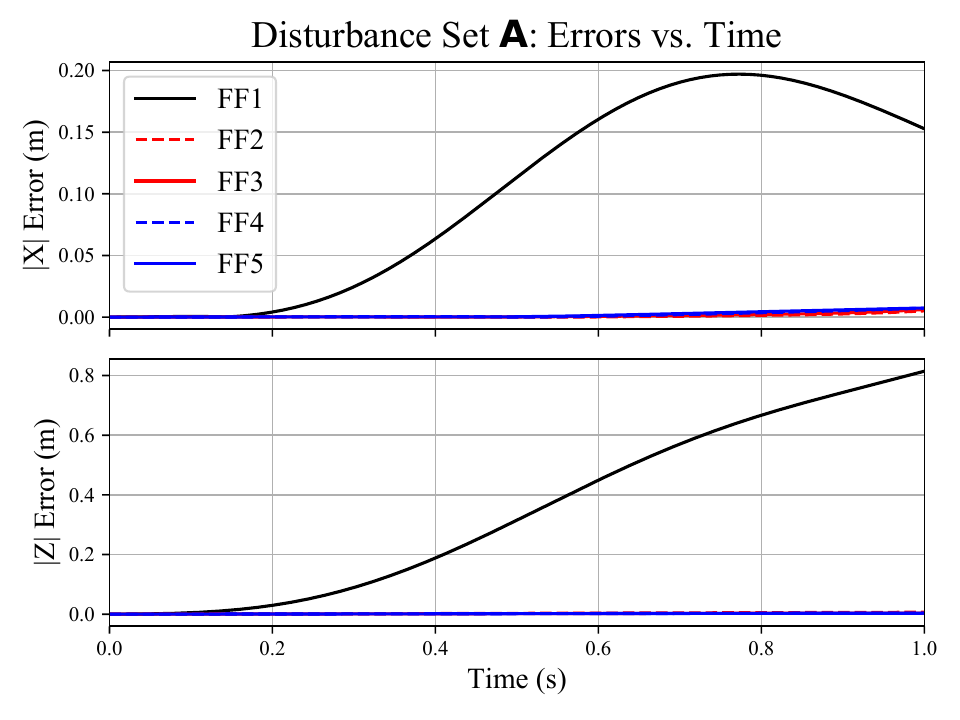}}
  \subfloat[Errors for disturbance set \textbf{B}, containing dynamic, but input-independent disturbances. \textbf{FF4} and \textbf{FF5}, which compensate for dynamic disturbances, perform the best.\label{fig:res_b}]{\includegraphics[width=8.3cm]{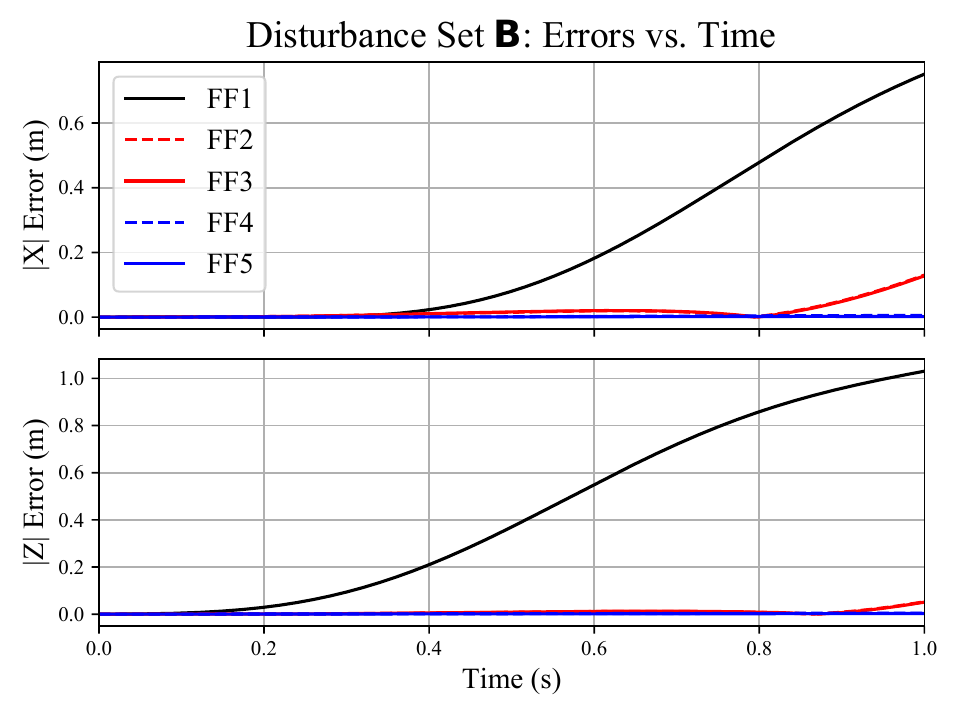}} \\
  \subfloat[Errors for disturbance set \textbf{C}, containing dynamic and input-dependent disturbances. Only \textbf{FF5} performs well.\label{fig:res_c}]{\includegraphics[width=8.3cm]{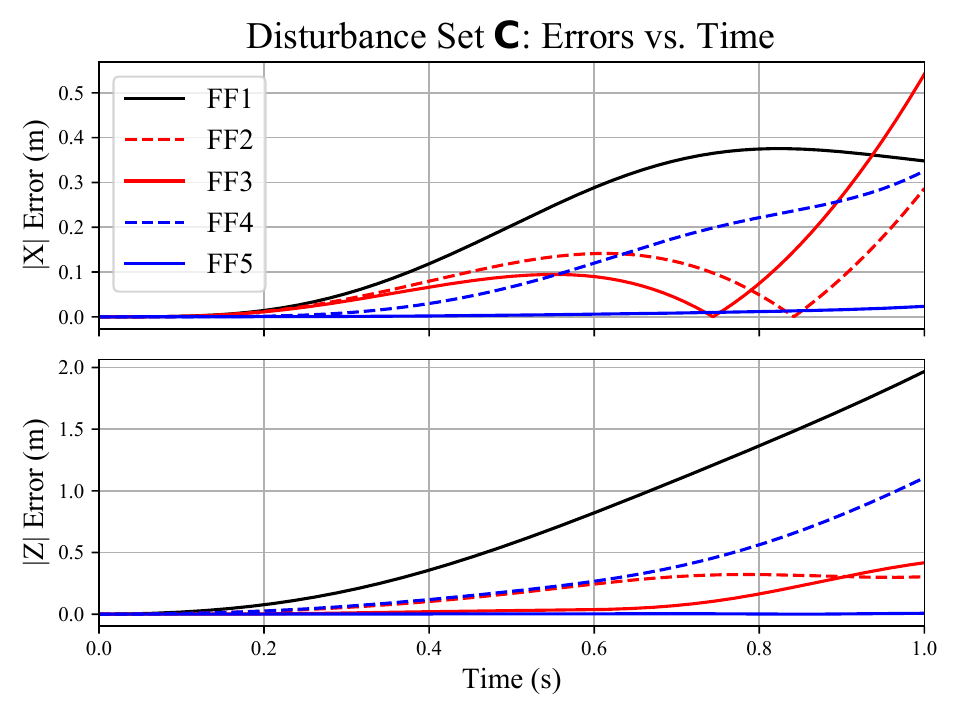}}
  \subfloat[Errors for disturbance set \textbf{D}, which contains all considered disturbances. \textbf{FF5} performs the best.\label{fig:res_d}]{\includegraphics[width=8.3cm]{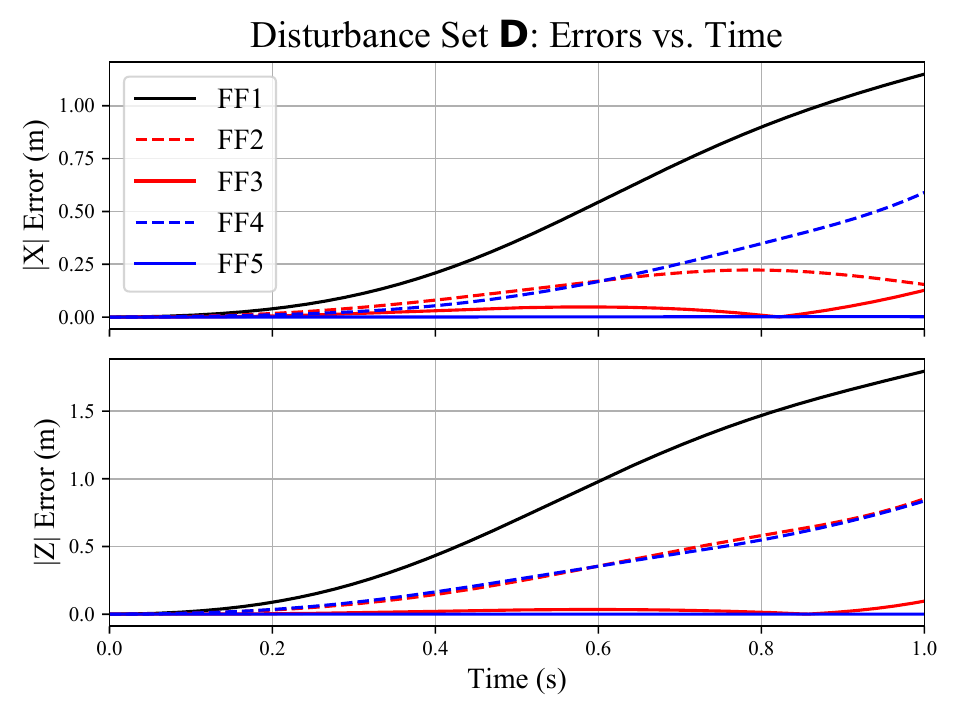}}
  \endgroup
  \caption{Error plots for all five feedforward strategies without feedback control under each of the four disturbance sets.}
  \label{fig:res}
\end{figure*}

Error plots for each of the four disturbance sets without feedback control are shown in Figs. \ref{fig:res_a} -- \ref{fig:res_d}. Under only a constant disturbance (Fig.~\ref{fig:res_a}), all disturbance compensation strategies work well, since the disturbance is neither input-dependent nor dynamic. When we introduce drag, a dynamic disturbance, in disturbance set \textbf{B} (Fig.~\ref{fig:res_b}), only the approaches that compensate for disturbance dynamics, \textbf{FF4} and \textbf{FF5}, achieve low error. Although basic disturbance compensation as in \textbf{FF2} helps considerably, accounting for disturbance dynamics improves performance further. Since in disturbance set \textbf{B}, the disturbances are still input-independent, the use of numerical optimization to solve the acceleration model \eqref{eq:accel_model} has no effect.

Under input-dependent disturbances, we see that \textbf{FF5} is the only approach that achieves low error. This is expected, as for both disturbance sets \textbf{C} and \textbf{D}, there are dynamic and input-dependent disturbances present.

\begin{figure}
  \centering
  \includegraphics[width=8.3cm]{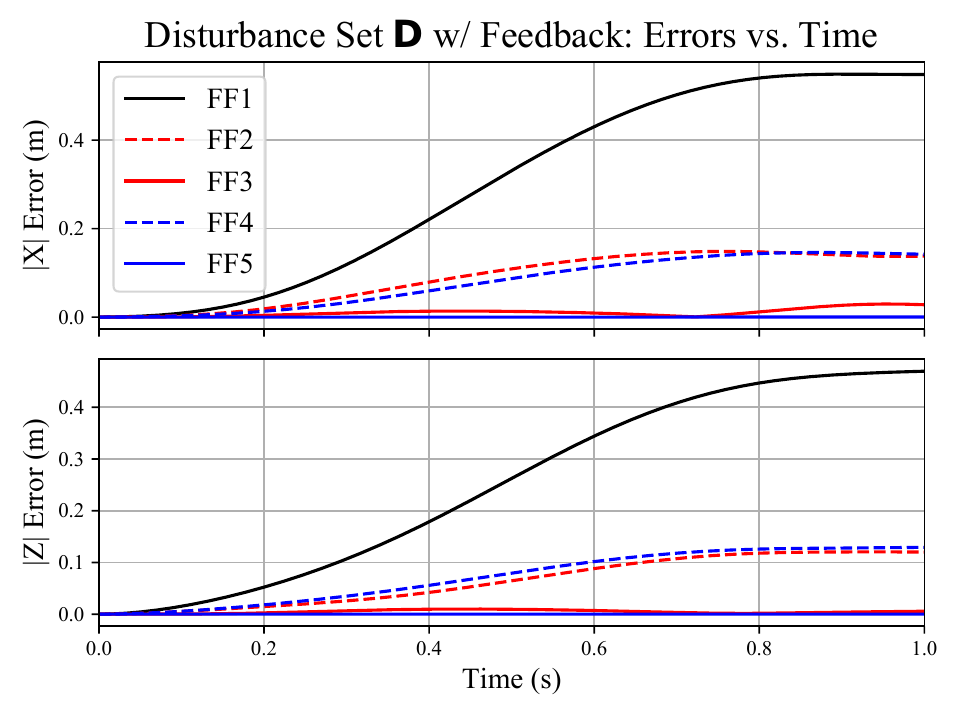}
  \caption{Error plots for all five feedforward strategies \textit{with} feedback control under disturbance set \textbf{D}. \textbf{FF5} outperforms all others.}
  \label{fig:res_d_fb}
\end{figure}

Error plots for disturbance set \textbf{D} \textit{with} feedback control are shown in Fig.~\ref{fig:res_d_fb}. We see that although feedback can reduce the error, it is not enough to completely eliminate the error. \textbf{FF5} still outperforms the other methods, achieving nearly zero error in all trials.

The maximum absolute position errors over the trajectory for all tested configurations are listed in Tables \ref{tab:results_nofb} and \ref{tab:results_fb}.

\begin{table}
  \caption{Maximum absolute position errors, in meters, for each control strategy without feedback in simulation}
  \label{tab:results_nofb}
  \begin{center}
  \begin{tabular}{c|c|c|c|c|c}
    \backslashbox{\textbf{Dist. Set}}{\textbf{Strategy}} & \textbf{FF1} & \textbf{FF2} & \textbf{FF3} & \textbf{FF4} & \textbf{FF5} \\
    \hline \textbf{A} & $0.829$ & $\mathbf{0.008}$ & $\mathbf{0.007}$ & $\mathbf{0.008}$ & $\mathbf{0.008}$ \\
    \hline \textbf{B} & $1.275$ & $0.140$ & $0.137$ & $\mathbf{0.007}$ & $\mathbf{0.003}$ \\
    \hline \textbf{C} & $1.998$ & $0.417$ & $0.684$ & $1.153$ & $\mathbf{0.025}$ \\
    \hline \textbf{D} & $2.131$ & $0.866$ & $0.159$ & $1.023$ & $\mathbf{0.002}$ \\
  \end{tabular}
  \end{center}
\end{table}

\begin{table}
  \caption{Maximum absolute position errors, in meters, for each control strategy with feedback in simulation}
  \label{tab:results_fb}
  \begin{center}
  \begin{tabular}{c|c|c|c|c|c}
    \backslashbox{\textbf{Dist. Set}}{\textbf{Strategy}} & \textbf{FF1} & \textbf{FF2} & \textbf{FF3} & \textbf{FF4} & \textbf{FF5} \\
    \hline \textbf{A} & $0.256$ & $\mathbf{0.001}$ & $\mathbf{0.001}$ & $\mathbf{0.001}$ & $\mathbf{0.001}$ \\
    \hline \textbf{B} & $0.412$ & $0.041$ & $0.042$ & $\mathbf{0.001}$ & $\mathbf{0.000}$ \\
    \hline \textbf{C} & $0.293$ & $0.069$ & $0.064$ & $0.076$ & $\mathbf{0.001}$ \\
    \hline \textbf{D} & $0.722$ & $0.189$ & $0.030$ & $0.194$ & $\mathbf{0.000}$ \\
  \end{tabular}
  \end{center}
\end{table}

\subsection{Hardware}

\begin{figure}
  \begin{center}
    \includegraphics[width=7cm]{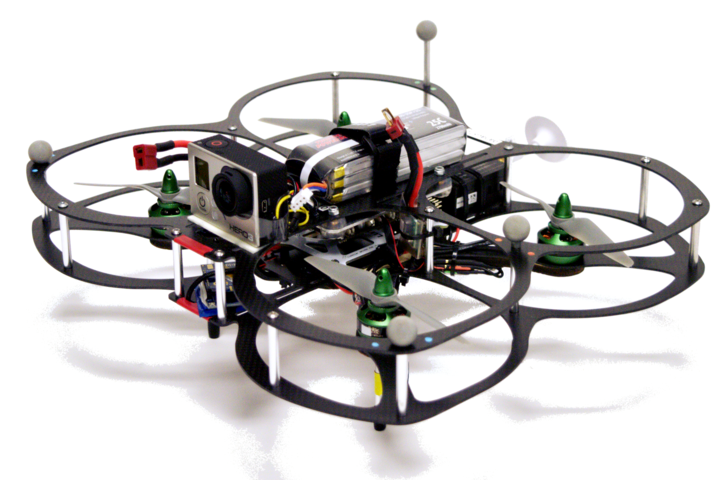}
  \end{center}
  \caption{The $750$ g quadrotor used for the hardware experiments. Onboard computation is performed by an Odroid XU4 and the Pixhawk 1 Flight Controller.}
  \label{fig:quadrotor}
\end{figure}

\subsubsection{Platform \& Setup}

To validate the usefulness of dynamic disturbance compensation and input-dependent disturbance compensation, we compare the five aforementioned feedforward generation strategies, \textbf{FF1} through \textbf{FF5}, on a $750$ g quadrotor while following aggressive trajectories.
Figure~\ref{fig:quadrotor} shows the hardware platform and Fig.~\ref{fig:overlay} shows the robot while following aggressive circle and line trajectories.

Position, velocity, and yaw feedback is provided by a motion capture arena at $100$ Hz, while pitch, roll, and angular velocity feedback is provided by a Pixhawk PX4 at $250$ Hz. Feedback control is performed by a cascaded PD system following \cite{mellinger}. The feedforward terms are as computed by \textbf{FF1} through \textbf{FF5}. \textbf{FF3} and \textbf{FF5} solve \eqref{eq:accel_de} numerically using the Newton-Raphson method. All control computation is performed onboard the vehicle's Odroid XU4 computer. The position control loop runs at $100$ Hz and the attitude control loop runs at $200$ Hz.

For the hardware experiments, we use three test trajectories: a $1.8$ s straight line trajectory, a circle trajectory, and a figure 8 trajectory. The trajectories are designed to be near the limit of what the robot can feasibly track. Table~\ref{tab:hardware_trajectories} lists the three trajectories and their maximum absolute derivatives.

\begin{table}[h]
\caption{The aggressive trajectories used to evaluate the proposed approach and their maximum derivatives.}
\label{tab:hardware_trajectories}
\begin{center}
\begin{tabular}{|c|c|c|c|c|c|}
  \hline \textbf{Traj.} & \textbf{$x$ (m)} & \textbf{$\dot x$ (m/s)} & \textbf{$\ddot x$ (m/s$^2$)} & \textbf{$x^{(3)}$ (m/s$^3$)} & \textbf{$x^{(4)}$ (m/s$^4$)} \\
  \hline Line & $2.7$ & $3.28$ & $6.26$ & $24.31$ & $216$ \\
  \hline Circle & $2.0$ & $2.75$ & $7.56$ & $21.43$ & $64.35$ \\
  \hline Figure 8 & $2.0$ & $2.75$ & $7.15$ & $21.43$ & $59.25$ \\
  \hline
\end{tabular}
\end{center}
\end{table}

\subsubsection{Model Learning}

We use linear regression as the model learning strategy in the hardware experiment.
Input data to the regression is a 6 dimensional vector consisting of the vehicle velocity and the commanded acceleration vector $\bm u$.
The system starts with an uninitialized model and uses a few test trajectories per trial to regress to the acceleration error.
The error model is then held fixed during the remaining trajectories used for error evaluation.
Although in principle, the model can be updated incrementally, keeping it fixed allows for a fair comparison between the control strategies.

\subsubsection{Results}

\begin{figure}
  \begin{center}
    \includegraphics[width=9.3cm]{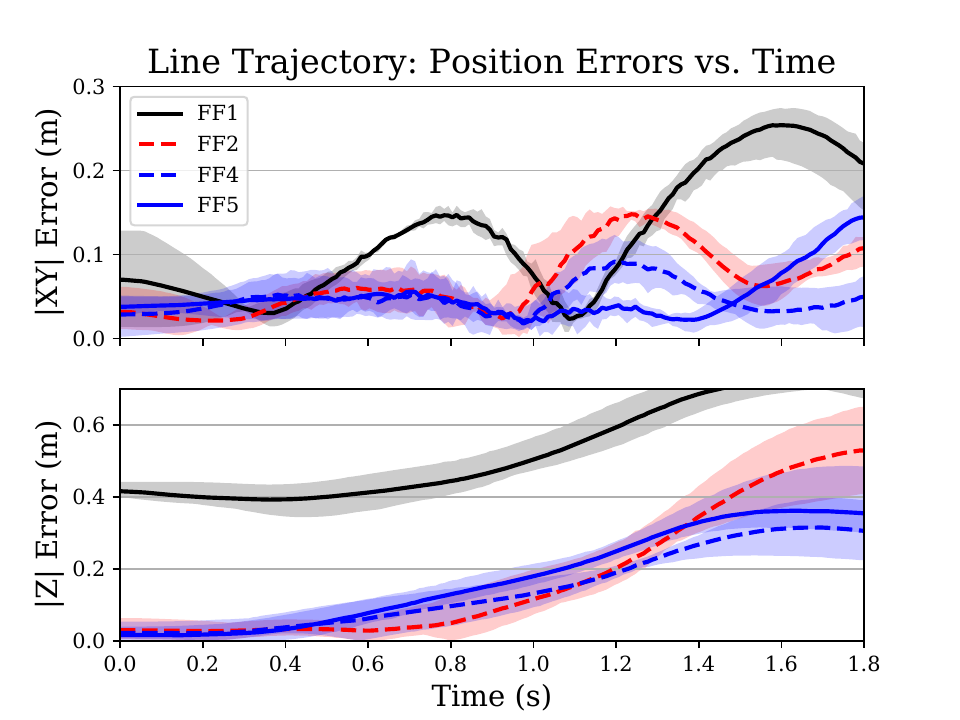}
  \end{center}
  \caption{Average absolute errors during an aggressive straight line trajectory for four of the five control strategies.
    Shaded regions denote the minimum and maximum errors per timestep over four trials.
    Means (m) ($\pm$ std (m)) over the 4 trajectories of the average $|x-y|$ error for \textbf{FF1}, \textbf{FF2}, \textbf{FF4}, and \textbf{FF5} respectively are 0.120 $\pm$ 0.003, 0.067 $\pm$ 0.003, \textbf{0.047} $\pm$ 0.009, and \textbf{0.048} $\pm$ 0.10.
    Those for the average $|z|$ error are 0.525 $\pm$ 0.023, 0.173 $\pm$ 0.034, \textbf{0.142} $\pm$ 0.025, and 0.173 $\pm$ 0.030.
  }
  \label{fig:res_line}
\end{figure}

For the line trajectory, each of \textbf{FF1}, \textbf{FF2}, \textbf{FF4}, and \textbf{FF5} is evaluated four times.
The first four trajectories, run using \textbf{FF1}, are used to train the acceleration error model.
An overlay of the vehicle executing the $2.7$ m line trajectory can be seen in Fig.~\ref{fig:overlay}.
Absolute errors along the trajectory and errors along the vertical axis for the line trajectory are shown in Fig.~\ref{fig:res_line}.
\textbf{FF1} performs the worst, especially along the vertical axis, indicating that the robot is underestimating the control input required to maintain hover.
\textbf{FF2} eliminates much of the error in the vertical axis, but still accumulates significant error along the trajectory, rising above $10$ cm consistently.
\textbf{FF4} and \textbf{FF5} provide on average a 30\% reduction in the average absolute $x-y$ tracking error along the trajectory when compared to \textbf{FF2}. This indicates that taking disturbance dynamics into account can significantly improve tracking performance.
This trajectory does not provide sufficient clarity to determine the impact of \textbf{FF5}, input-dependent disturbance compensation.

\begin{figure}
  \begin{center}
    \includegraphics[width=9.3cm]{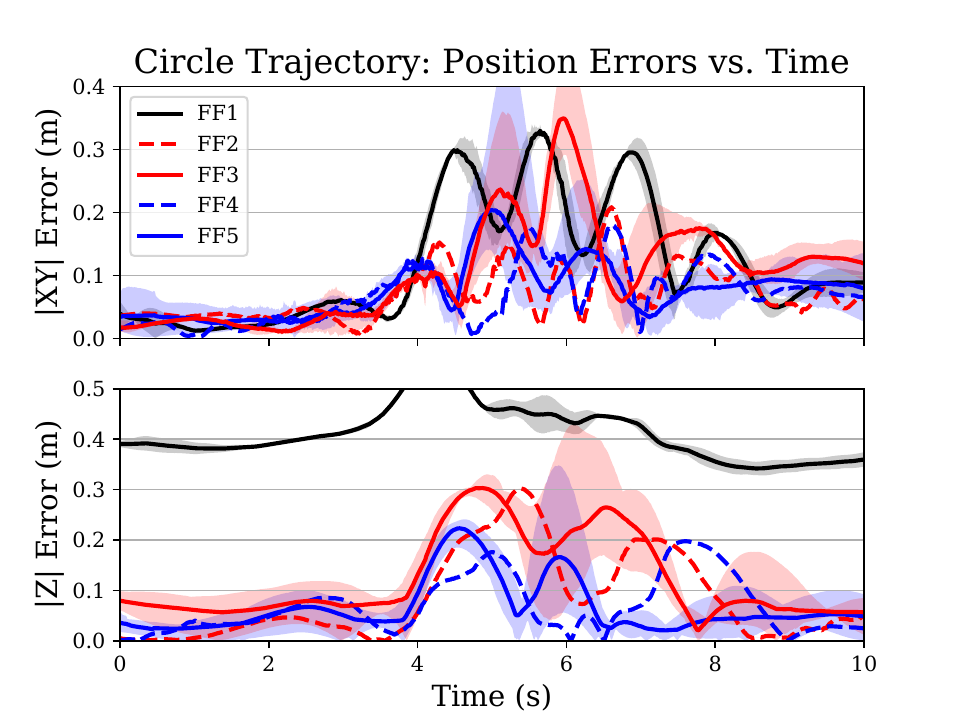}
  \end{center}
  \caption{Average absolute errors during an aggressive circle trajectory for the five control strategies. Shaded regions denote the minimum and maximum errors per timestep.
    Avg. $|x-y|$ errors (m) for \textbf{FF1}, \textbf{FF2}, \textbf{FF3}, \textbf{FF4}, and \textbf{FF5} are 0.118, \textbf{0.071}, 0.103, \textbf{0.069}, and \textbf{0.076}, respectively, while avg. $|z|$ errors (m) are 0.41, 0.084, 0.122, \textbf{0.069}, and \textbf{0.066}, respectively.
  }
  \label{fig:res_circle}
\end{figure}

For the circle trajectory, all of the feedforward strategies are evaluated once, with \textbf{FF3} and \textbf{FF5} receiving two and four more trajectories respectively.
An overlay of the vehicle executing the circle trajectory can be seen in Fig.~\ref{fig:overlay}.
Fig.~\ref{fig:res_circle} shows the resulting error.
As expected \textbf{FF1}, with no disturbance compensation, performs the worst. \textbf{FF2}, \textbf{FF4}, and \textbf{FF5} all perform similarly well, with \textbf{FF2} achieving slightly lower vertical error than the others.
\textbf{FF3} performs slightly worse here than \textbf{FF2}, suggesting that the numerical routine may be failing to converge or that the input's dependence on the acceleration error has not been properly modeled.

\begin{figure}
  \begin{center}
    \includegraphics[width=9.3cm]{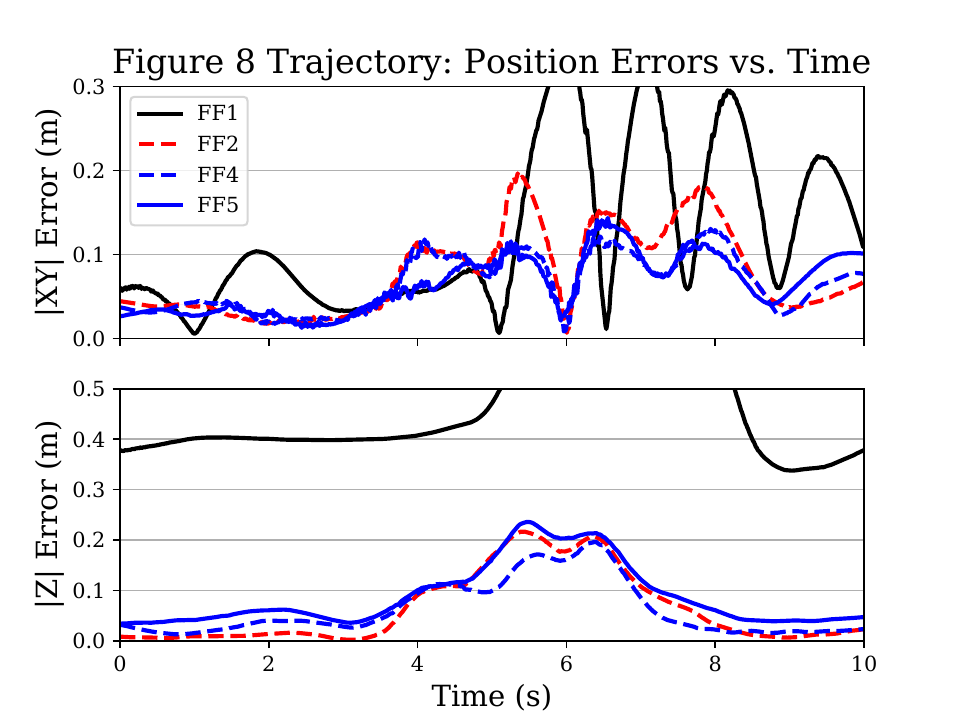}
  \end{center}
  \caption{Errors during an aggressive figure 8 trajectory for four of the five control strategies.
    Avg. $|x-y|$ errors (m) for \textbf{FF1}, \textbf{FF2}, \textbf{FF4}, and \textbf{FF5} are 0.105, 0.076, \textbf{0.063}, and \textbf{0.059} respectively,
    while avg. $|z|$ errors (m) are 0.473, \textbf{0.064}, \textbf{0.064}, and 0.088, respectively.
  }
  \label{fig:res_figure8}
\end{figure}

Fig.~\ref{fig:res_figure8} shows the error of \textbf{FF1}, \textbf{FF2}, \textbf{FF4}, and \textbf{FF5} along the figure 8 trajectory for one trial each. The improvement of \textbf{FF4} over \textbf{FF2} here is smaller than in the other trajectories, suggesting that dynamic disturbances have relatively less of an impact when following the figure 8, though more experimental trials are warranted to strengthen this claim.

\section{CONCLUSION}

We have presented a method that allows compensation of dynamic disturbances through evaluation of the derivatives of a learned model. We have shown in both simulation and hardware experiments that our dynamic disturbance compensation method improves performance over traditional disturbance compensation. We have also shown the usefulness of input-dependent disturbance compensation in simulation and preliminary results on hardware. The versatility of the approach in a realistic robotics application has been verified through evaluation on three distinct test trajectories.

Future work will evaluate nonlinear regression techniques, such as ISSGPR, on hardware platforms, as well as consider regression techniques that explicitly optimize model derivative accuracy.
An interesting avenue of future study is to analyze theoretically how the error model accuracy affects the performance of each of the feedforward generation strategies.
Lastly, we hope to apply this technique to the attitude dynamics of multirotors, in order to fully compensate for vehicle disturbances and modeling errors.

\section{ACKNOWLEDGMENTS}

The authors thank Xuning Yang for helpful feedback on this manuscript.

\bibliographystyle{plainnat}
\bibliography{refs,applications}

\end{document}